\documentclass[sigconf]{aamas}  

\AtBeginDocument{%
  \providecommand\BibTeX{{%
    \normalfont B\kern-0.5em{\scshape i\kern-0.25em b}\kern-0.8em\TeX}}}
\usepackage{booktabs}

\usepackage{flushend}
\setcopyright{ifaamas}  
\copyrightyear{2020} 
\acmYear{2020} 
\acmDOI{} 
\acmPrice{} 
\acmISBN{} 
\acmConference[AAMAS'20]{Proc.\@ of the 19th International Conference on Autonomous Agents and Multiagent Systems (AAMAS 2020)}{May 9--13, 2020}{Auckland, New Zealand}{B.~An, N.~Yorke-Smith, A.~El~Fallah~Seghrouchni, G.~Sukthankar (eds.)}  

\usepackage{subfig}
\usepackage{multirow}
\usepackage{multicol}
\usepackage{graphicx}
\usepackage{url}
\usepackage{microtype}
\usepackage{mathtools}
\usepackage{float}
\floatstyle{plaintop}
\restylefloat{table}
\usepackage{amsmath}
\usepackage{amssymb}  
\usepackage{mathtools}
\usepackage{fancyhdr}
\usepackage{kantlipsum}
\usepackage{eso-pic}
\usepackage{nomencl}
\usepackage{etoolbox}
\usepackage{balance}
\usepackage{dsfont}
\usepackage{tikz}
\usepackage{eucal}
\usepackage{upgreek}
\tikzset{>=latex}
\usetikzlibrary{fit,automata,positioning,angles,quotes}
\usepackage[algo2e,linesnumbered,ruled,lined]{algorithm2e}
\usepackage{booktabs}
\usepackage{hyperref}
\hypersetup{
	colorlinks=true,
	linkcolor=black,
	filecolor=black,      
	urlcolor=black,
	citecolor=black,}

\newtheorem{remark}{Remark}[section]
\newtheorem{problem}{\textbf{Problem}}

\DeclareMathOperator*{\argmax}{argmax}

\begin{document}

\title[Cautious Reinforcement Learning with Logical Constraints]{Cautious Reinforcement Learning with Logical Constraints}
\author{Mohammadhosein Hasanbeig}
\affiliation{\institution{University of Oxford}
\city{Oxford}\country{UK}}
\email{hosein.hasanbeig@cs.ox.ac.uk}

\author{Alessandro Abate}
\affiliation{\institution{University of Oxford}
\city{Oxford}\country{UK}}
\email{alessandro.abate@cs.ox.ac.uk}

\author{Daniel Kroening}
\affiliation{\institution{University of Oxford}
\city{Oxford}\country{UK}}
\email{kroening@cs.ox.ac.uk}

\renewcommand{\shortauthors}{Hasanbeig, Abate and Kroening}


\begin{abstract}
This paper presents the concept of an adaptive \emph{safe padding} that forces Reinforcement Learning (RL) to synthesise optimal control policies while ensuring safety during the learning process. 
Policies are synthesised to satisfy a goal, expressed as a temporal logic formula, with maximal probability. Enforcing the RL agent to stay safe during learning might limit the exploration,  
however we show that the proposed architecture is able to automatically handle the trade-off between efficient progress in exploration (towards goal satisfaction) and ensuring safety. 
Theoretical guarantees are available on the optimality of the synthesised policies and on the convergence of the learning algorithm. Experimental results are provided to showcase the performance of the proposed method. 
\end{abstract}

\maketitle

\section{Introduction}

Reinforcement Learning (RL) is an algorithm with no supervision that can be used to train an agent to interact with an unknown environment. The dynamics of the interaction are often assumed to be a Markov Decision Process (MDP). The key feature of RL is its sole dependence on a set of experiences, which are gathered by interacting with, i.e., \emph{exploring}, the environment. 
This makes RL inherently different than classical dynamic programming methods~\cite{puterman} and automatic control approaches, in the sense that it can optimally solve the decision-making problem without any prior knowledge about the MDP model~\cite{sutton}. 
This very practical feature has paved the way for applications of RL in economics, engineering, and biology \emph{inter alia}, to solve sequential decision-making problems when no model is available~\cite{ng,ifac,chemistry,lcnfq,deepql,cdc,silver}. 

While the existing RL methods deliver good training outcomes, by and large they lack guarantees on what happens \emph{during training}. Existing results rely either on ``soft safety'' or on ``ergodicity'' assumptions. The essence of soft safety is that unsafe states, which are to be avoided, may still be visited regardless of the consequent catastrophic outcome. The ergodicity assumption means that any state is eventually reachable from any other state if a proper policy is followed -- this second assumption allows a (non-episodic) RL agent to explore by simply favouring states that have rarely been visited, even if they are unsafe in practice. 
These assumptions might be reasonable for certain applications, such as gaming or virtual environments, but are not affordable for many safety-critical physical systems, which may break before exploration completes. Thus, unsurprisingly, most of the existing exploration methods are impractical in safety-critical scenarios where the aforementioned assumptions do not hold.

\emph{Safe RL} is an active area of research focusing on the training of agents with guarantees on safety~\cite{garcia}. 
However, most of these methods minimize the risk that the \emph{trained agent} violates a safety specification, but do not ensure safety of  exploration \emph{during training}~\cite{risk1,risk2,risk3,risk4,arxiv,shield,shield2,lyap,lcrl_j}. Recent approaches on this problem~\cite{entropy,barrier_1,barrier_2,knownD,knownD2} are either computationally expensive or require explicit, strong assumptions about the model of agent-environment interactions.  

In this work we take a step back from currently used exploration methods and recall that RL is originally inspired by cognitive and behavioural psychology~\cite{psy}. When humans learn to control, they naturally account for what \emph{they expect} to be safe, i.e., (1)~they use their own a-priori prediction of the environment when choosing which behaviours to explore, and (2)~they continuously update their knowledge and expectations using local observations. For example, when novice pilots learn to control a helicopter, they slowly pull the lever until the helicopter slightly lifts off the ground, then quickly land it back down. They will repeat this a few times, gradually increasing the time the helicopter hovers off the ground. At all times, they aim to prevent the likelihood of a disaster by ensuring that a safe landing is possible. In other words, they try to restrict exploration to a locally safe state-action region, which in this work we shall name \emph{safe padding}. As their knowledge of the dynamics of the helicopter in its environment improves, they perform increasingly more sophisticated maneuvers, namely exploring new sequences of state-action pairs. It is interesting to notice that the maneuvers that a fully trained pilot will ultimately perform are initially incompatible with the safety of the learning agent, and as such might be located outside of the \emph{safe padding} initially in use while learning. 

Inspired by the cognitive approach to learning outlined above, we propose to equip the RL agent with a limited knowledge of its own dynamics, and with its local perception of safety. \emph{Uncertain dynamics} characterise how the environment reacts to the actions of the agent. Much like a trainee pilot, the agent starts by performing exploratory cautious actions, and gradually, in line with the growing confidence about the environment obtained from observations, the range of acceptably safe actions grows, and the uncertain component of the dynamics becomes known. 

Beyond the issue of safe exploration, in the RL literature task satisfaction is usually achieved by hand-engineering appropriate rewards~\cite{garcia}. In this context the difficulty is mapping complex, possibly long-term, sequential tasks to an appropriate reward structure~\cite{reduction}. If done incorrectly, 
the outcome of learning might be unexpectedly sub-optimal.  
As an extension of ongoing research \cite{lcrl_j,cdc}, 
in this work we employ temporal logic, and more specifically Linear Temporal Logic (LTL)~\cite{pnueli} as a formal reward shaping technique to specify task-related goals~\cite{clarke}. 
We convert a given LTL formula into an automaton that expresses the property~\cite{bible}, then translate the automaton into a state-adaptive reward structure. 
Using any off-the-shelf RL with the obtained reward structure results in policies that maximised the probability of verifying the given LTL formula. 
This framework, which we call Logically-Constrained RL (LCRL \cite{lcrl_j,cdc}), is enhanced here by safe learning, with an architecture named \emph{cautious RL}. Cautious RL is applicable to any standard reward-based RL. 

While safety could be as well one of the tasks expressed in the LTL formula,  meaning that the given LTL property will hold when deploying the trained agent,  safety in the context of this work will be separately accounted for during training by means of safe padding.  
Technically, we propose a safe padding in combination with the state-adaptive reward function based on the task automaton over the state-action pairs of the MDP. Using this automatic reward shaping procedure, RL is able to generate a policy that satisfies the given task expressed as an LTL property with maximal probability, while the safe padding prevents violating safety during learning. Thus, the method we propose inherits aspects of reward engineering that are standard in RL, and at the same time infuses notions from formal methods that allow guiding the exploration safely, furthermore also certifying the learning outcomes in terms of the probability of satisfying the given task expressed as an LTL formula.


The proposed framework is related to, but cannot be reduced to, work on Constrained MDP (CMDP)~\cite{cmdp}, due to its generality and to its inherent structural differences: in this work LTL satisfaction is encoded into the expected return itself, while in CMDP algorithms the original objective is separate from the constraint. 
Focusing exclusively on the safety fragment of LTL, the concept of shielding is proposed in~\cite{shield}: the proposed shield is a reactive machine that ensures that the agent remains safe during learning. To express the specification, \cite{shield} uses a DFA and then translates the problem into a safety game. This work has been extended to probabilistic CTL properties in~\cite{shield2}, where a probabilistic model checking technique is used to construct the shield. Unlike this paper, in both \cite{shield,shield2}, the agent has to observe the entire MDP (and opponents) to construct a model of the safety game. 
\cite{fulton2,fulton} address safety-critical settings in the context of cyber-physical systems, where the agent has to select a correct model within a heterogeneous set of models in model-based RL: \cite{fulton} first generates a set of feasible models given an initial model and data on runs of the system. With such a set of feasible models, the agent has to learn how to safely identify which model is the most accurate; \cite{fulton3} further employs differential dynamic logic~\cite{ddl}, a first-order logic for specifying and proving properties of hybrid models. 

In summary, in this paper we contribute the following:
\begin{enumerate}
    \item Cautious RL: a safe exploration scheme for model-free RL. This is applicable to standard reward-based RL, however, we tailor it to the next goals: 
    \item The use of LTL as task specification for policy synthesis in RL. Automatic reward shaping and task decomposition when the task is highly complex. Bringing (1) and (2) together, we obtain: 
    \item Prediction of unsafe state-action pairs (safe padding) while learning and consequent limitation of exploration and of policy learning for LTL task satisfaction. The method guarantees asymptotic results.      
\end{enumerate} 


\section{Background}\label{background}
\subsection{Problem Setup}
\begin{definition} [Markov Decision Process, MDP]\label{def_mdp} 
	A finite MDP $\mathfrak{M}$ is a six tuple $(\allowbreak
	\mathcal{S},\allowbreak\mathcal{A},\allowbreak s_0,\allowbreak
	P,\allowbreak\mathcal{AP},\allowbreak L)$ where $\mathcal{S}$ is a finite set called the state space, $\mathcal{A}$ is a finite set of actions, $s_0$
	is the initial state. $P(\cdot|s,a)\in\mathcal{P}(\mathcal{S})$ is the probability distribution over the next states given that action $a$ has been taken in state $s$, where $\mathcal{P}(\mathcal{S})$ is the set of probability distributions on subsets of $\mathcal{S}$. 
	$\mathcal{AP}$ is a finite set of atomic propositions and a labelling
	function $L: \mathcal{S} \rightarrow 2^{\mathcal{AP}}$ assigns to each state
	$s \in \mathcal{S}$ a set of atomic propositions $L(s) \subseteq
	2^\mathcal{AP}$. $\hfill\square$
\end{definition}
A random variable $R(s,a)\sim\rho(\cdot|s,a)\in\mathcal{P}(\mathbb{R^+})$ can be defined over the MDP $\mathfrak{M}$, representing the immediate reward obtained when action $a$ is taken in a given state $s$ where $\mathcal{P}(\mathbb{R^+})$ is the set of probability distributions on subsets of $\mathbb{R^+}$, and $\rho$ is the reward distribution. One possible realization of the immediate reward is denoted by $r(s,a)$.
\begin{definition}[Stationary Deterministic Policy]
    A policy is a rule according to which the agent chooses its action at a given state. More formally, a policy $\pi$ is a mapping from the state space to a distribution in $\mathcal{P}(\mathcal{A})$, where $\mathcal{P}(\mathcal{A})$ is the set of probability distributions on subsets of $\mathcal{A}$. A policy is stationary if $\pi(\cdot|s)\in\mathcal{P}(\mathcal{A})$ does not change over time and it is called a deterministic policy if $\pi(\cdot|s)$ is a degenerate distribution. $\hfill\square$
\end{definition}
An MDP controlled by a policy $\pi$ induces a Markov chain $\mathfrak{M}^\pi$ with transition kernel $P^\pi(\cdot|s)=P(\cdot|s,\pi(s))$, and with reward distribution $\rho^\pi(\cdot|s)=\rho(\cdot|s,\pi(s))$ such that $R^\pi(s)\sim\rho^\pi(\cdot|s)$.
\begin{definition}
	[Expected Discounted Return \cite{sutton}] \label{expectedreturn} For any policy~$\pi$ on an MDP $\mathfrak{M}$, the expected discounted return in state $s$ is
	\begin{equation}
	    {V}^\pi_\mathfrak{M}(s)=\mathds{E}^\pi[\sum\limits_{n=0}^{\infty} \gamma^n~ r(s_n,a_n)|s_0=s],   
	\end{equation}
	where $\mathds{E}^\pi[\cdot]$ denotes the expected value by following policy $\pi$, $\gamma$~is the discount factor, and $s_0,a_0,s_1,a_1...$ is the sequence of state-action pairs generated by policy $\pi$. The expected discounted return is often referred to as \emph{value function}. Note that the discount factor~$\gamma$ is a hyper-parameter that has to be tuned. In particular, there is standard work in RL on state-dependent discount factors \cite{discount,discount2,discount3}, which is shown to preserve convergence and optimality guarantees. Similarly, the action-value function is defined as:
	\begin{equation}\label{return}
	    {Q}^\pi_\mathfrak{M}(s,a)=\mathds{E}^\pi[\sum\limits_{n=0}^{\infty} \gamma^n~ r(s_n,a_n)|s_0=s,a_0=a].
	\end{equation}
	We drop the subscript $\mathfrak{M}$ when it is clear from the context. $\hfill\square$
\end{definition}
\begin{definition}[Optimal Policy]\label{optimal_pol}
	An optimal policy $\pi^*$ is defined as follows:
	$$
		\pi^*(s)=\arg\sup\limits_{\pi \in \varpi}~ {V}^\pi_\mathfrak{M}(s),
	$$
	where $\varpi$ is the set of stationary deterministic policies over  $\mathcal{S}$. $\hfill\square$
\end{definition}

\begin{theorem}[\cite{puterman,puterman2}]\label{optimal_policy_thm}
	In an MDP $\mathfrak{M}$ with a bounded reward function and a finite action space optimal policies are stationary and deterministic.
\end{theorem}
By Theorem \ref{optimal_policy_thm}, as long as the reward function is bounded and the action space is finite, the optimal policy in Definition \ref{optimal_pol} exists. This would be the case for this work. 

\subsection{Linear Temporal Logic}
LTL formulae over a given set of atomic propositions $\mathcal{AP}$ are syntactically defined as \cite{pnueli}
\begin{equation}\label{ltlsyntax}
	\varphi::= \mathit{true} ~|~ \alpha ~|~ \varphi \land \varphi ~|~ \neg \varphi ~|~ \bigcirc \varphi ~|~ \varphi ~\mathrm{U}~ \varphi,
\end{equation}
where $\alpha\in \mathcal{AP}$, and the operators $ \bigcirc $ and $ \mathrm{U} $ are called \emph{next} and \emph{until}, respectively. Using the until operator we define two further temporal modalities: 
(1)~eventually, $\lozenge \varphi = \mathit{true} ~\mathrm{U}~ \varphi$; and 
(2)~always, $\square \varphi = \neg \lozenge \neg \varphi$. An infinite \emph{word} $w$ over the alphabet $2^{\mathcal{AP}}$ in MDP $\mathfrak{M}$ is defined as an infinite sequence $w=l_0 ~l_1 ~l_2 ~l_3 ...\in (2^{\mathcal{AP}})^{\omega}$, where $\omega$ denotes infinite repetition and $l_i\in2^{\mathcal{AP}}$, $\forall i\in\mathbb{N}$. The language $\{w \in (2^{\mathcal{AP}})^\omega ~\mbox{s.t.}~ w \models \varphi\}$ is defined as the set of words that satisfy the LTL formula $\phi$, where $\models\subseteq (2^{\mathcal{AP}})^{\omega}\times\phi$ is the satisfaction relation.

\begin{definition}[Probability of Satisfying an LTL Formula] \label{ltlprobab}
Starting from any state $s$ and following a stationary deterministic policy $\pi$, we denote the probability of satisfying formula $\varphi$ as	
$
\mathds{P}(s..^{\mathit{\pi}} \models \varphi),
$
where $s..^{\pi}$ is the collection of all state sequences starting from $ s $, generated under policy $\pi$. $\hfill\square$
\end{definition}

For any LTL property $\varphi$ the set $\mathit{Words}(\varphi)$ can be expressed by an LDBA. An LDBA is a special form of a Generalized B\"uchi Automaton (GBA) \cite{sickert}, defined as follows: 
\begin{definition}[Generalized B\"uchi Automaton] 
	A GBA $\mathfrak{A}=(\allowbreak\mathcal{Q},\allowbreak q_0, \allowbreak\Delta,\allowbreak\Sigma, \allowbreak\mathcal{F})$ is a structure where $\mathcal{Q}$ is its finite set of states, $q_0 \in \mathcal{Q}$ is the initial state, $\Delta: \mathcal{Q} \times \Sigma \rightarrow 2^\mathcal{Q}$ is a transition relation, $\Sigma=2^{\mathcal{AP}}$ is a finite alphabet, and $\mathcal{F}=\{\mathcal{F}_1,...,\mathcal{F}_f\}$ is the set of accepting conditions, where $\mathcal{F}_j \subseteq \mathcal{Q}, 1\leq j\leq f$.~$\hfill\square$ 
\end{definition}

Let $\Sigma^\omega$ be the set of all infinite words over $\Sigma$. An infinite word $w \in \Sigma^\omega$ is accepted by a GBA $\mathfrak{A}$ if there exists an infinite run $\theta \in\mathcal{Q}^\omega$ starting from $q_0$ where $\theta[i+1] \in\Delta(\theta[i],w[i]),~i \geq 0$ and for each $\mathcal{F}_j \in \mathcal{F}$ 
\begin{equation} \label{acc}
	\mathit{inf}(\theta) \cap \mathcal{F}_j \neq \emptyset,
\end{equation}
where $\mathit{inf}(\theta)$ is the set of states that are visited infinitely often by the run $\theta$. 

\begin{definition}[LDBA]\label{ldbadef}
	A GBA $\mathfrak{A}=(\allowbreak\mathcal{Q},\allowbreak q_0, \allowbreak\Delta,\allowbreak\Sigma, \allowbreak\mathcal{F})$ is limit-determ\-inistic if $\mathcal{Q}$ can be partitioned into two disjoint sets $\mathcal{Q}=\mathcal{Q}_N \cup \mathcal{Q}_D$ such that~\cite{sickert}:
	\begin{itemize}
		\item $\Delta(q,\alpha) \subset \mathcal{Q}_D$ and $|\Delta(q,\alpha)|=1$ for every state $q\in\mathcal{Q}_D$ and for every $\alpha \in \Sigma$;
		\item for every $\mathcal{F}_j \in \mathcal{F}$, $\mathcal{F}_j \subseteq \mathcal{Q}_D$; and
		\item $q_0 \in \mathcal{Q}_N$, and all the transitions from $\mathcal{Q}_N$ to $\mathcal{Q}_D$ are non-deterministic $\varepsilon$-transitions\footnote{An $\varepsilon$-transition allows an automaton to change its state without reading any label.}.~$\hfill\square$ 
	\end{itemize}
\end{definition}
\begin{remark}\label{ldba_remark}
    An LDBA is a GBA that has two partitions: one initial ($\mathcal{Q}_N$) and one accepting ($\mathcal{Q}_D$). The accepting part includes all the accepting states and has deterministic transitions. The LTL-to-LDBA construction used in this paper \cite{sickert} results in an automaton with deterministic initial and accepting parts. According to Definition~\ref{ldbadef}, the discussed structure is still an LDBA, though the determinism in the initial part is stronger than that required in the LDBA definition. We explain later why this matters for the proposed algorithm.
\end{remark}

\begin{remark}\label{nonblocking}
    At any state $q$ of an LDBA $\mathfrak{A}$, the output of the transition relation $\Delta$ is non-empty, namely all the states of $\mathfrak{A}$ are non-blocking. Further, any subset of $\Sigma$ can be read at any state.~$\hfill\square$
\end{remark}

\section{Cautious Reinforcement Learning with Logical Constraints}\label{lcsec} 
\subsection{Logically-guided Reinforcement Learning}

In order to relate the structure of an  MDP to that of an LDBA, for now we assume that the MDP graph and its transition probabilities are fully known. This allows us to formally define a synchronised structure that will be key for policy synthesis. This  assumption is dropped entirely later, and we stress that policy synthesis can indeed be implemented on-the-fly over unknown MDPs via model-free RL. 

\begin{definition} [Product MDP]\label{product_mdp_def}
	Given an MDP $\mathfrak{M}=(\allowbreak
	\mathcal{S},\allowbreak\mathcal{A},\allowbreak s_0,\allowbreak
	P,\allowbreak\mathcal{AP},\allowbreak L)$ and an LDBA $\mathfrak{A}=(\allowbreak\mathcal{Q},\allowbreak q_0, \allowbreak\Delta,\allowbreak\Sigma, \allowbreak\mathcal{F})$ with $\Sigma=2^{\mathcal{AP}}$, the product MDP is defined as $\mathfrak{M}\otimes\mathfrak{A}= \mathfrak{P}=(\mathcal{S}^\otimes,\allowbreak \mathcal{A}^\otimes,\allowbreak s^\otimes_0,P^\otimes,\allowbreak \mathcal{AP}^\otimes,\allowbreak L^\otimes,\allowbreak \mathcal{F}^\otimes)$, where $\mathcal{S}^\otimes = \mathcal{S}\times\mathcal{Q}$, $s^\otimes_0=(s_0,q_0)$, $\mathcal{AP}^\otimes = \mathcal{Q}$, $L^\otimes : \mathcal{S}^\otimes\rightarrow 2^\mathcal{Q}$ such that $L^\otimes(s,q)={q}$ and $\mathcal{F}^\otimes \subseteq {\mathcal{S}^\otimes}$ is the set of accepting states $\mathcal{F}^\otimes=\{\mathcal{F}^\otimes_1,...,\mathcal{F}^\otimes_f\}$, where $\mathcal{F}^\otimes_j=\mathcal{S}\times \mathcal{F}_j$. The transition kernel $P^\otimes(\cdot|s_i^\otimes,a)\in\mathcal{P}(\mathcal{S}^\otimes)$ is such that given the current state $(s_i,q_i)$ and action $a$, the new state $(s_j,q_j)$ is obtained such that $s_j\sim P(\cdot|s_i,a)$ and $q_j\in\Delta(q_i,L(s_j))$. In order to handle $\varepsilon$-transitions in $\mathfrak{A}$ we furthermore need to add the following transitions:
	\begin{itemize}
		\item for every potential $\varepsilon$-transition to some state $q \in \mathcal{Q}$ we add a corresponding action $\varepsilon_q$ in the product:
		$$
			\mathcal{A}^\otimes=\mathcal{A}\cup \{\varepsilon_q, q \in \mathcal{Q}\}.
		$$
		
		\item The transition probabilities corresponding to $\varepsilon$-transitions are given by 
		\[P^\otimes((s_j,q_j)|(s_i,q_i),\varepsilon_q) = \left\{
		\begin{array}{lr}
			1 & s_i=s_j, q_i\xrightarrow{\varepsilon_q} q_j=q,\\
			0 & \text{otherwise}. 
		\end{array}
		\right.
		\]
	\end{itemize}
\end{definition}
\vspace{-4mm}
$\hfill\square$\\*

\setlength{\fboxrule}{0pt}
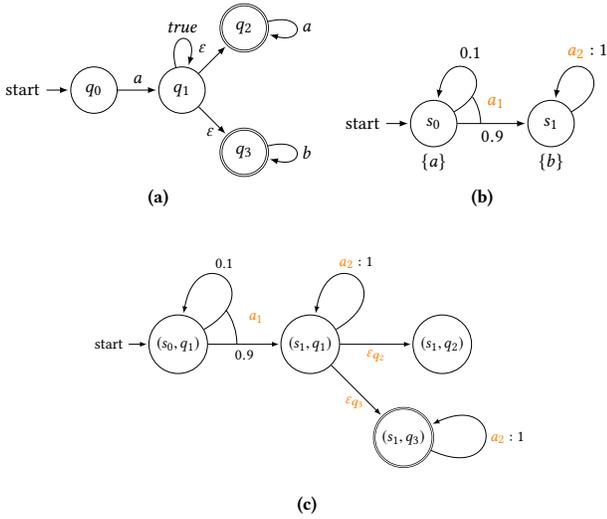
\begin{figure}[!t]\centering
\subfloat[]{{
    \scalebox{.78}{
		\begin{tikzpicture}[shorten >=1pt,node distance=1.5cm,on grid,auto] 
		\node[state,initial] (q_0)   {$q_0$}; 
		\node[state] (q_1) [right=of q_0]  {$q_1$}; 
		\node[state,accepting] (q_2) [above right=of q_1] {$q_2$}; 
		\node[state,accepting] (q_3) [below right=of q_1] {$q_3$};
		\path[->] 
		(q_0) edge node {$a$} (q_1)
		(q_1) edge  node {$\varepsilon$} (q_2)
		(q_1) edge [loop above] node {$\mathit{true}$} (q_1)
		(q_1) edge [below] node {$\varepsilon~~~$} (q_3)
		(q_2) edge  [loop right] node {$a$} (q_2)
		(q_3) edge  [loop right] node {$b$} (q_3);
		\end{tikzpicture}
		}
		}}
\subfloat[]{{
	\scalebox{.78}{
		\begin{tikzpicture}[shorten >=1pt,node distance=2cm,on grid,auto] \node[state,initial,label=below:$\{a\}$] (s_0) {$s_0$};
		\node[state,label=below:$\{b\}$] (s_1) [right=of s_0]{$s_1$};
		\draw[->] (s_0) -- node [below] {$0.9$} (s_1);
		\draw[->] (s_0) [out=30,in=80,loop] to coordinate[pos=0.2](aa) node [above] {$\fbox{0.1}$} (s_0);
		\draw[->] (s_1) [out=30,in=80,loop] to coordinate[pos=0.2] node [above] {\fbox{$\textcolor{orange}{a_2}:1$}} (s_1);
		\path pic[draw, angle radius=8mm,"$\textcolor{orange}{a_1}$",angle eccentricity=1.4] {angle = s_1--s_0--aa};
		\end{tikzpicture}
		}
		}}\\
\subfloat[]{
		\scalebox{.65}{
		\begin{tikzpicture}[shorten >=1pt,node distance=2.7cm,on grid,auto] 
		\node[state,initial] (q_0)   {$(s_0,q_1)$}; 
		\node[state] (q_1) [right=of q_0]  {$(s_1,q_1)$}; 
		\node[state] (q_2) [right=of q_1] {$(s_1,q_2)$}; 
		\node[state,accepting] (q_3) [below right=of q_1] {$(s_1,q_3)$};
		\draw[->] (q_0) -- node [below] {$0.9$} (q_1);
		\draw[->] (q_0) [out=30,in=80,loop] to coordinate[pos=0.2](aa) node [above] {$\fbox{0.1}$} (q_0);
		\draw[->] (q_1) [out=30,in=80,loop] to coordinate[pos=0.2] node [above] {\fbox{$\textcolor{orange}{a_2}:1$}} (q_1);
		\draw[->] (q_1) -- node [below] {$\textcolor{orange}{\varepsilon_{q_2}}$} (q_2);
		\draw[->] (q_1) -- node [below] {$\textcolor{orange}{\varepsilon_{q_3}}~~$} (q_3);
		\path pic[draw, angle radius=12mm,"$\textcolor{orange}{a_1}$",angle eccentricity=1.4] {angle = q_1--q_0--aa};
		\draw[->] (q_3) [out=335,in=25,loop] to coordinate[pos=0.2] node [right] {$\textcolor{orange}{a_2}:1$} (q_3);
		\end{tikzpicture}}
	}\vspace*{-0.25cm}
	\caption{Example of product MDP: (a) the LDBA for $\varphi=a\wedge\bigcirc(\lozenge\square a\vee\lozenge\square b)$, (b) an instance MDP, and (c) the product according to Def. \ref{product_mdp_def}.}
	\label{fig:product_mdp_ex}
\end{figure}

An example of a product MDP, 
as per Definition~\ref{product_mdp_def}, 
is given in Fig.~\ref{fig:product_mdp_ex} for an instance of an MDP and for an LDBA generated from  the LTL formula 
$$
\varphi=a\wedge\bigcirc(\lozenge\square a\vee\lozenge\square b).
$$
Next, we propose a state-adaptive reward function based on the accepting condition of the automaton, so that maximisation of the expected cumulative reward (to be attained via RL) implies the  maximisation of the satisfaction probability for the LTL formula (Definition \ref{ltlprobab}). 

\subsection{State-adaptive Reward}

Before introducing the state-adaptive reward shaping scheme 
we need to provide a few definitions. 

\begin{definition}[Non-accepting Sink Component]
	A non-accepting sink component of an automaton, in this case an LDBA, $\mathfrak{A}=(\allowbreak\mathcal{Q},\allowbreak q_0, \allowbreak\Delta,\allowbreak\Sigma, \allowbreak\mathcal{F})$ is a directed graph induced by a set of states $ \mathcal{Q}_\mathit{sink} \subset\mathcal{Q}$ such that (1) the graph is strongly connected; (2) it does not include all accepting sets $ \mathcal{F}_k,~k=1,...,f $; and (3) there exists no other strongly connected set $ \mathcal{Q}_\mathit{sink}' \subset \mathcal{Q},~\mathcal{Q}_\mathit{sink}'\neq \mathcal{Q}_\mathit{sink} $ such that $ \mathcal{Q}_\mathit{sink} \subset \mathcal{Q}_\mathit{sink}' $. We~denote the union of all non-accepting sink components as $\mathcal{Q}_\mathit{sinks}$.~$\hfill\square$ 
\end{definition}
\begin{remark}\label{unsafeness}
    Note that after taking a transition in the automaton that takes us to $\mathcal{Q}_\mathit{sinks}$, it is not possible to satisfy the associated LTL property anymore, namely the probability of LTL satisfaction becomes zero under any policy.  Identifying $\mathcal{Q}_\mathit{sinks}$ allows the agent to predict immediate labels that lead to a violation of the property. Thus, transitions $q \xrightarrow{\alpha\in\Sigma} q'$ to $\mathcal{Q}_\mathit{sinks}$ in the automaton are denoted by  $\Delta_\mathit{sinks}$.~\hfill $\square$
\end{remark}
\begin{definition}
	[Accepting Frontier Function]\label{frontier} Given an LDBA $\mathfrak{A} \allowbreak =(\allowbreak\mathcal{Q},\allowbreak q_0, \allowbreak\Delta,\allowbreak\Sigma, \allowbreak\mathcal{F})$, we define the accepting frontier function $ AF:\mathcal{Q}\times 2^{\mathcal{Q}}\rightarrow2^\mathcal{Q} $, which executes the following operation over any given set $ \mathds{F}\in 2^{\mathcal{Q}}$: 
	\begin{equation}\label{acc_function}
	AF(q,\mathds{F})=\left\{
	\begin{array}{l@{\hspace{0.2cm}:\hspace{0.2cm}}l}
	\mathds{F}_{\setminus \mathcal{F}_j}~~~ & (q \in \mathcal{F}_j) \wedge (\mathds{F}\neq \mathcal{F}_j)\\
	\bigcup\limits_{k=1}^{f}{\mathcal{F}_k} _{\setminus \mathcal{F}_j} & (q \in \mathcal{F}_j) \wedge (\mathds{F}=\mathcal{F}_j).
	\end{array}
	\right.
	\end{equation}
\end{definition}
\vspace{-4mm}
$\hfill\square$\\*
Now assume that the agent is at state $ s^\otimes=(s,q) $, takes action $ a $ and observes the subsequent state $ {s^\otimes}'=(s',q') $. Note that since both the initial and accepting parts of the LDBA are deterministic, $q'$ can be obtained on-the-fly\footnote{There is no need to \emph{explicitly build} the product MDP and to store all its states in memory. The automaton transitions can be executed on-the-fly, as the agent reads the labels of the MDP states.}. The immediate reward is a scalar value, determined according to the following rule:  
\begin{equation}\label{thereward}
	\begin{aligned}
		R(s^\otimes,a) = \left\{
		\begin{array}{lr}
			r_p & $ if $  q' \in \mathds{A},~{s^\otimes}'=(s',q'),\\
			r_n & \text{otherwise}.
		\end{array}
		\right.
	\end{aligned}
\end{equation} 
Here, $r_p>0$ is a positive reward and $r_n=0$ is a neutral reward. The set $ \mathds{A} \in 2^{\mathcal{Q}}$ is called the \emph{accepting frontier set}, and it is initialized as the family set of all accepting sets, i.e., 
\begin{equation}\label{eq:initA}
    \mathds{A}=\{F_k\}_{k=1}^{f}.
\end{equation}

The accepting frontier set is updated on-the-fly every time a set $\mathcal{F}_j$ is visited as $\mathds{A}\leftarrow AF(q',\mathds{A})$ 
where $AF(q',\mathds{A})$ is the accepting frontier function defined before.

In short, after initialisation of $\mathds{A}=\{F_k\}_{k=1}^{f}$ the accepting frontier function $AF$ always excludes from $\mathds{A}$ those accepting sets that have been visited or are being visited, unless it is the only remaining accepting set. In this case the accepting frontier $ \mathds{A} $ is reset, as per the second condition of \eqref{acc_function}. Thus, intuitively the set $ \mathds{A} $ always contains those accepting states that ought to be visited at any given time: in this sense the reward function is adapted to the accepting condition from the LDBA. The agent is guided by the above reward assignment to visit the accepting sets infinitely often, and consequently, to satisfy the given LTL property~$\varphi$,  as per \eqref{acc}.   
As in~\cite{lcrl_j,cdc}, we will argue that we can learn an optimal policy generating traces that satisfy the given property~$\varphi$ with maximum probability. 

\begin{remark}
    As in Definition \ref{ldbadef} and Remark \ref{ldba_remark}, the automaton transitions can be executed by reading the labels \emph{only}, which makes the agent aware of the automaton state without explicitly constructing the product MDP. The transitions in the automaton can be executed on-the-fly as the agent reads the labels of the MDP states, without knowledge of the model structure or the transition probabilities (or their product). As such, our algorithm is implementable in a fully model free manner. ~\hfill $\square$
\end{remark}

\subsection{Safe Padding for Exploration}

Ensuring safe exploration is critical when RL is employed to generate control policies in situations when learning from failure is unacceptable or extremely expensive, as in safety-critical autonomy for instance. We call this problem \emph{safe policy synthesis}.   
We propose a \emph{safe padding} for the agent by leveraging the agent limited knowledge about its own dynamics and its local perception of safety. Hence, the agent avoids violating the safety requirement (up to some probability level), while learning the optimal policy for task satisfaction.  

\begin{problem}[Safe Policy Synthesis]\label{problem_def}  
Given an unknown black-box MDP $\mathfrak{M}$ and an LTL property~$\varphi$ a learning agent attains the following: (1) synthesises an optimal policy $\pi^*$ via RL such that the induced Markov chain $\mathfrak{M}^{\pi^*}$ satisfies the LTL property with maximum possible probability; and (2) does not violate a safety requirement during learning. We assume that the transition probability function $P$ in the MDP is only partly known. Technically, we assume that (i) the agent acquires prior knowledge about its own transition kernel (dynamics) $P_a:\mathcal{S}\times\mathcal{A}\times\mathcal{S}\rightarrow [0,1]$, that might not be accurate in the environment $\mathfrak{M}$ in which the agent operates. We also assume that (ii) the agent has a limited observability \emph{only} of the labelling function $L$ in Definition~\ref{def_mdp}: without knowing the full structure of the MDP, the agent is able to observe the labels of the surrounding states up to some distance from the current position.~$\hfill\square$
\end{problem}

Let us assume a starting belief about the agent transition kernel $P_a$, to be encoded as Dirichlet distributions 
\cite{pac_littman,mle} via two functions $ \Psi: \mathcal{S} \times \mathcal{A} \rightarrow \mathds{N}$ and $ \psi: \mathcal{S} \times \mathcal{A} \times \mathcal{S} \rightarrow \mathds{N} $. Function $ \psi(s,a,{s'}) $ represents the number of times the agent executes action $ a $ in state~$s$, thereafter moving to state $ {s}' $, and $ \Psi(s,a)=\sum_{{s}' \in \mathcal{S}} \psi(s,a,{s}')$. The function $ \Psi(s,a) $ is initialised to be one for every state-action pair, reflecting the fact that at any given state it is possible to take any action, and also avoiding division by zero; the function $ \psi(s,a,{s'}) $ is initialised to zero. Once the transition $(s,a,{s'})$ is taken for the first time, $\Psi(s,a)~\leftarrow~2$, so $\psi(s,a,{s'})$ has to be incremented to $2$ to reflect the correct belief ${P}_a(s,a,{s'})=1$ (Algorithm~\ref{alg:sparl}, lines 17--23). 

\medskip

The \emph{safe padding} is a subset of the state-action space of the MDP that the agent considers safe to explore. 
As the agent explores and learns, the safe padding slowly expands, as much as in the flight control example the pilots slowly expands their comfort zone to learn how to control the helicopter. The expansion rate of the safe padding depends on how many times a particular state-action pair has been visited and on the corresponding update of the transition kernel~$P_a$. The expansion mechanism and kernel update are explained shortly. 

Let us recall that we have assumed that the agent has a limited observation range over the labels of the surrounding states (as per Problem \ref{problem_def}). Assume that the observation radius is $r_o\geq1$, meaning that the agent can \emph{only} see the labels of states that have distance at most $r_o$ over the MDP graph, i.e.
$$
O(s)=\{s' \in \mathcal{S}~\wedge~d(s,s')\leq r_o\},
$$
where $O(s)\subset\mathcal{S}$ is the set of states whose labels are visible at $s$, i.e., the agent current state, and $d(s,s')$ is the length of the shortest directed path from $s$ to $s'$. Note that the agent can only see state labels,  
and has to rely on its current knowledge of the dynamics, namely $P_a$. Let the current state of the agent in the product MDP be $s^\otimes=(s,q)$. Define 
\begin{equation}
    O_\mathit{safe}(s)=\{x \in O(s),~q\xrightarrow{L(x)}q'\notin\Delta_\mathit{sinks}\},
\end{equation}
as the set of safe states. Given the observation radius and the current belief of the agent about its own transition dynamics $P_a$, the agent performs a local, finite-horizon Bellman update over the observation area $O(s)$. For each state $x\in O_\mathit{safe}(s)$ and a horizon $H$ the Bellman update is performed $H$ times as follows~\cite{NDP}: 
\begin{align}\label{uk}
\begin{aligned}
& u_k(x)=\min\limits_{a\in\mathcal{A}}\sum_{{x}'} P_a(x,a,{x}') u_{k+1}({x}'),~k=H-1,...,0\\
& u_H(x)=\mathds{1}_{O_\mathit{safe}}(x),
\end{aligned}
\end{align}
where $x'$ is the subsequent state after taking action $a$ in $x$, $u_k:\mathcal{S}\rightarrow[0,1]$ is a local value function at time step $k$, and $H$ is initialized as $H=r_o$. The value $u_H(x)$ is initialised as $1$ if $x \in O_\mathit{safe}(s)$, and as $0$ otherwise. 
With this initialisation, $u_0$ represents the agent estimation of the minimum probability of staying in $O_\mathit{safe}$ within $H$ steps, i.e. $u_0=\mathds{P}_{\min}(\square^{\leq H}~ O_{safe})$. Notice that this estimation is indeed pessimistic and conservative. 
Hence, with a fixed horizon $H$ the agent is able to calculate the maximum probability of violating the LTL property by picking action $a$ at state $s^\otimes$ in $H$ steps:
\begin{equation}\label{uk2}
    U_H(s^\otimes,a)=1-\sum_{{s}'} P_a(s,a,{s}') u_{0}({s}'),
\end{equation}
where $s^\otimes=(s,q)$. As assumed in Problem~\ref{problem_def}, the agent dynamics $P_a$ is then updated considering the Maximum Likelihood Estimation (MLE) for the mean ${P}_a(s,a,{s'})$~\cite{pac_littman,mle} as:
\begin{equation}\label{uk3}
{P}_a(s,a,{s'})\leftarrow\dfrac{\psi(s,a,{s}')}{\Psi(s,a)}.
\end{equation}
Here, $ \psi(s,a,{s'}) $ represents the number of times the agent executes action $ a $ in state $ s $, thereafter moving to state $ {s}' $, whereas $ \Psi(s,a)=\sum_{{s}' \in \mathcal{S}} \psi(s,a,{s}')$. Note that $\psi$ and $\Psi$ (and consequently $P_a$) are functions of the MDP state and action spaces, not of the product MDP, since they capture the agent dynamics over the original MDP, which remains the same regardless of the current automaton state $q$. Hence, the RHS of \eqref{uk2} only depends on state $s$ and action~$a$, and not the automaton state~$q$.  

\begin{remark}
    Note that for each state $s$, the Bellman updates in \eqref{uk} are  performed only over $O_\mathit{safe}(s)$. Recall that the set $O_\mathit{safe}(s)$ is the safe subset within the bounded observation area and thus, the computational burden of \eqref{uk} is limited.~\hfill $\square$
\end{remark}

\subsection{Safe Policy Synthesis with Logical Constraints}

At this point we bring together the use of the safe padding with the constrained learning architecture generating policies that satisfy the given LTL formula. 
In order to pick the most optimal yet safe action at each state, we propose a \emph{double learner} architecture, as  explained in the following. 

The first part is an \emph{optimistic} learner that employs Q-learning (QL) \cite{watkins} to maximize the expected cumulative return as in \eqref{return}. For each state $s^\otimes \in \mathcal{S}^\otimes$ and for any action $a \in \mathcal{A}^\otimes$, QL assigns a quantitative value $Q:\mathcal{S}^\otimes\times\mathcal{A}^\otimes\rightarrow \mathbb{R}^+$, which is initialized with an arbitrary and finite value over all state-action pairs. The Q-function is updated by the following rule when the agent takes action $ a $ at state $ s^\otimes $:
\begin{equation}\label{ql_update_rule}
\scalebox{0.95}{$
    Q(s^\otimes,a) \leftarrow Q(s^\otimes,a)+\mu[R(s^\otimes,a)+\gamma \max\limits_{a' \in \mathcal{A}^\otimes}(Q({s^\otimes}',a'))-Q(s^\otimes,a)],$}
\end{equation}
where $ Q(s^\otimes,a) $ is the Q-value corresponding to state-action $ (s^\otimes,a) $, $ 0<\mu\leq 1 $ is called learning rate or step size, $ R(s^\otimes,a) $ is the reward obtained for performing action $a$ in state $s^\otimes$, $0\leq\gamma\leq 1$ is the discount factor, and ${s^\otimes}'$ is the state obtained after performing action~$a$. The Q-function for the rest of the state-action pairs remains unchanged. 

\begin{algorithm2e}[!t]
\DontPrintSemicolon
\SetKw{return}{return}
\SetKwRepeat{Do}{do}{while}
\SetKwFunction{terminate}{episode$\_$terminate}
\SetKwFor{terminatedef}{episode$\_$terminate()}{}{}
\SetKwData{conflict}{conflict}
\SetKwData{safe}{safe}
\SetKwData{sat}{sat}
\SetKwData{unsafe}{unsafe}
\SetKwData{unknown}{unknown}
\SetKwData{true}{true}
\SetKwData{false}{false}
\SetKwInOut{Input}{input}
\SetKwInOut{Output}{output}
\SetKwFor{Loop}{Loop}{}{}
\SetKw{KwNot}{not}
\begin{small}
	\Input{LTL specification, $ \textit{it\_threshold} $, $ \gamma $, $ \mu $, $r_o$, $p_\mathit{critical}$, $P_a$}
	\Output{$\pi^*$}
	convert the desired LTL property to an equivalent LDBA $\mathfrak{A}$\;
	initialize $\mathds{A}=\{F_k\}_{k=1}^{f}$\;
	initialize $\kappa=1,~\forall s \in\mathcal{S}$\;
	initialize horizon $H=r_o,~\forall s \in\mathcal{S}$\;
	initialize $Q: \mathcal{S}^\otimes \times \mathcal{A}^\otimes \rightarrow \mathbb{R}^+$\;
	initialize $\textit{episode-number}:=0$\;
	initialize $\textit{iteration-number}:=0$\;
	\While{$Q$ is not converged}
	{
		$\textit{episode-number}\leftarrow\textit{episode-number}+1$\;
		$s^\otimes=(s_0,q_0)$\;
		\While{$ (q \notin \mathcal{Q}_\mathit{sink}:~s^\otimes=(s,q))~ \wedge ~ (\textit{iteration-number}<\text{it\_threshold})$}
		{
			$\textit{iteration-number}\leftarrow\textit{iteration-number}+1$\;
			$\#$ \textbf{pessimistic learner}\;
			~~~calculate $U_H(s^\otimes,a)$ using $P_a$ as in \eqref{uk2}\;
			~~~generate $\mathcal{A}^H_p(s^\otimes)$ as in \eqref{eq:a_h} \;
			~~~choose $a_*=\argmax_{a \in\mathcal{A}_p^H[1:\kappa]}~Q(s^\otimes,a)- r_p U_H(s^\otimes,a)$ \;
			~~~$ \Psi(s^\otimes,a_*)\leftarrow \Psi(s^\otimes,a_*)+1$\;
			~~~execute action $a_*$ and observe the next state $s^\otimes_*$\;
			~~~if $\Psi(s^\otimes,a_*)=2$ then\;
			~~~\vrule~~~{	$\psi(s^\otimes,a_*,{s^\otimes_*})=2 $\;
			}
			~~~else\;
			~~~\vrule~~~{	$\psi(s^\otimes,a_*,{s^\otimes_*})\leftarrow\psi(s^\otimes,a_*,{s^\otimes_*})+1$\;
			}
			~~~end\;
			~~~update $P_a(s,a_*,{s_*})$ as in \eqref{uk3}\;
			~~~update $H(s)$\;
			~~~update $\kappa(s)$\;
			$\#$ \textbf{optimistic learner}\;
			~~~receive the reward $R({s^\otimes},a_*)$\;
			~~~$\mathds{A}\leftarrow \mathit{Acc}(s_*,\mathds{A})$\;
			~~~$Q({s^\otimes},a_*)\leftarrow Q({s^\otimes},a_*)+\mu[R({s^\otimes},a_*)-Q({s^\otimes},a_*)+\gamma \max_{a'}(Q(s^\otimes_*,a'))]$\;
			~~~$s^\otimes=s^\otimes_*$\;
		}
	}
\end{small}
\caption{Cautious RL}
\label{alg:sparl}
\end{algorithm2e}

Under mild assumptions over the learning rate, for finite-state and -action spaces QL converges to a unique limit, call it $Q^*$ \cite{watkins}. Once QL converges, the optimal policy $\pi^*: \mathcal{S}^\otimes \rightarrow \mathcal{A}^\otimes$ for $\mathfrak{P}$ can be generated by selecting the action that yields the highest $Q^*$, i.e.,
$$
	\pi^*(s^\otimes)=\arg\max\limits_{a \in \mathcal{A}^\otimes}~Q^*(s^\otimes,a). 
$$
It has been shown~\cite{lcrl_j,cdc} that the optimal policy $\pi^*$ generates traces that satisfy the given property~$\varphi$ with maximum probability.

Of course, adhering to the optimistic learner policy by no means guarantees to keep the agent safe \emph{during} the exploration. This is where the second part of the architecture, i.e., the \emph{pessimistic} learner, is needed: we exploit the concept of cautious learning and create a safe padding for the agent to explore safely. The pessimistic learner locally calculates $U_H(s^\otimes,a),\forall a \in \mathcal{A}^\otimes$ for a selected  horizon $H$ at the current state $s^\otimes$, and then outputs a set of permissive (safe) actions for the optimistic learner. Define a hyper-parameter $p_\mathit{critical}$ called \emph{critical probability} to select actions $a\in\mathcal{A}^\otimes$. This is the probability that is considered to be critically risky (unsafe) and is defined prior to learning: any action $a$ at state $s^\otimes$ that has $U_H(s^\otimes,a)\geq p_\mathit{critical}$ is considered as a critical action and has to be avoided. Accordingly, we introduce
\begin{equation}\label{eq:a_h}
    \mathcal{A}_p^H(s^\otimes)=\{a\in\mathcal{A}^\otimes:U_H(s^\otimes,a)<p_\mathit{critical}\}.
\end{equation}

This set is sorted over actions such that the first element has the lowest $U_H(s^\otimes,a)$ -- with slight abuse of notations we write $\mathcal{A}_p^H[k]$ for the $k$-th element. 

\begin{figure}[!t]
	\centering
	\vspace{-6mm}
	\scalebox{0.88}[1]{
	\hspace{-5.5mm}\subfloat[]{{\includegraphics[width=0.50\linewidth]{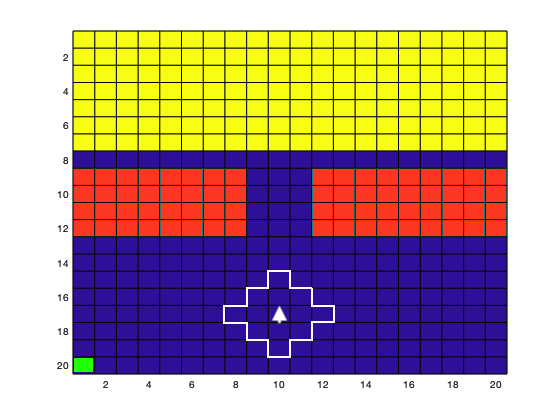}}}}%
	\subfloat[]{{\includegraphics[width=0.50\linewidth]{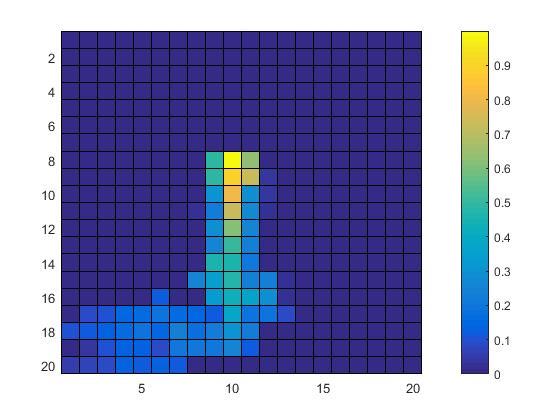}}}%
	\qquad
	\subfloat[]{{\includegraphics[width=0.40\linewidth]{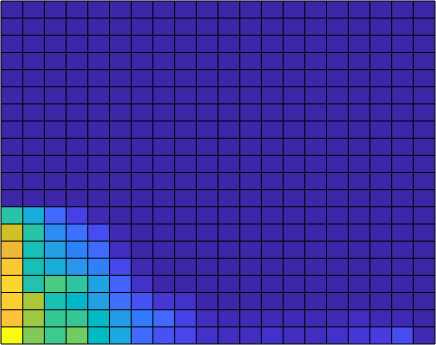}}}\hspace{7mm}
	\subfloat[]{{\includegraphics[width=0.40\linewidth]{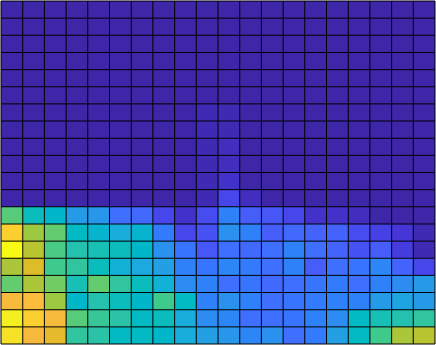}}}%
	\qquad
	\subfloat[]{{\includegraphics[width=0.55\linewidth]{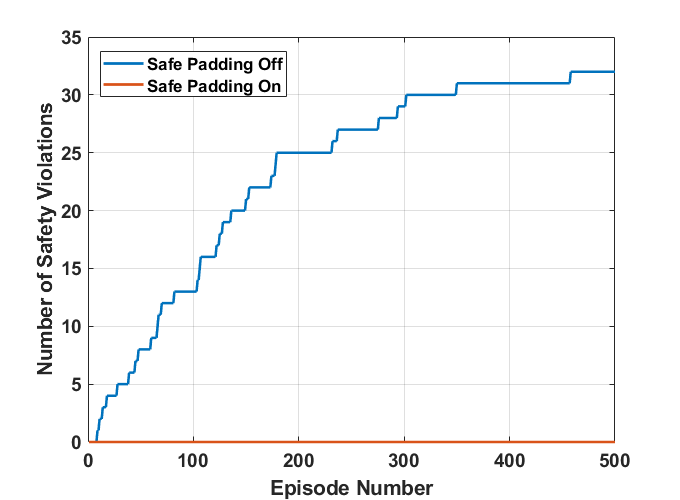}}}%
	\caption{(a) slippery grid world and agent, represented by an arrow surrounded by an observation area. Labelling is yellow: $\mathit{target}$, red: $\mathit{unsafe}$, blue: $\mathit{safe}$, and green is the initial state $s_0$; (b)~final value function $V(s)$, (c)--(d) state visitation number $v(s)=\sum_{a}\Psi(s,a)$ vs.~time where the safe padding gradually grows and repels the agent from entering unsafe region; (e) number of times the agent reaches unsafe area (red) until RL converges with safe padding on vs.~off.}%
	\label{fig_bridge}
\end{figure}

At the beginning the pessimistic learner is conservative and only allows those actions in $\mathcal{A}_p^H$ that have index of less than $\kappa$, i.e., $\mathcal{A}_p^H[1:\kappa]$, where $\kappa$ is a monotonically increasing function of the number of state visitations $v(s)=\sum_{a}\Psi(s,a)$, such that
$$\kappa(v)|_{v=1}=1 \hspace{2mm}\mbox{and}\hspace{2mm} \lim_{v\rightarrow\infty}\kappa(v)=|\mathcal{A}_p^H|.$$
The horizon $H$ follows the opposite rule, namely it is a monotonically decreasing function of $v(s)$ such that initially 
$$H(v)|_{v=1}=r_o \hspace{2mm}\mbox{and}\hspace{2mm} \lim_{v\rightarrow\infty} H(v)=1.$$
In other words, when the uncertainty around a state is high, the agent looks ahead as much as possible, i.e.~$H=r_o$. Once the confidence level around that particular state increases then the agent considers riskier decisions by just looking one step ahead, i.e.~$H=1$. This essentially means that the safe padding grows as the uncertainty diminishes (or learning grows). Note that in practice, $\kappa(v)$ and $H(v)$ can be step-wise functions of~$v$, and thus the agent is not necessarily required to visit a state an infinite number of times to get to $H=1$ and $\kappa=|\mathcal{A}_p^H|$. 

\begin{figure}[!t]
	\centering
	\vspace{-4mm}
	\scalebox{0.88}[1]{
	\hspace{-5.5mm}\subfloat[]{{\includegraphics[width=0.5\linewidth]{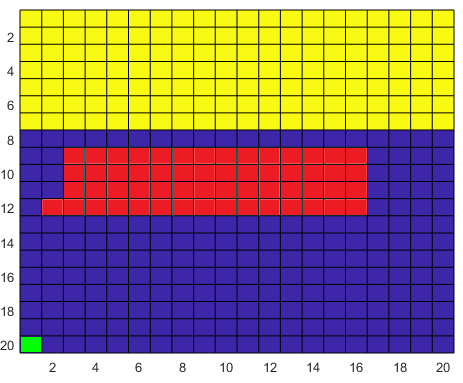}}}}\\
	\subfloat[]{{\includegraphics[width=0.50\linewidth]{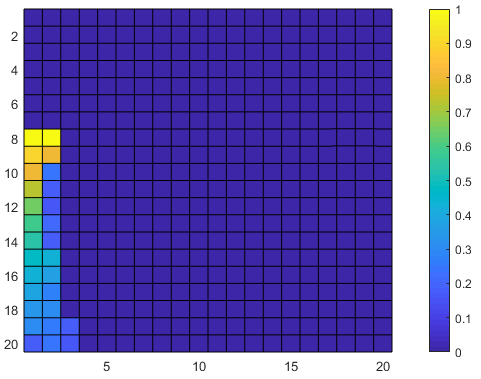}}}%
	\subfloat[]{{\includegraphics[width=0.50\linewidth]{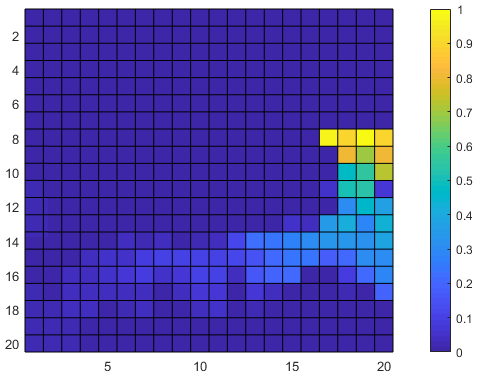}}}%
	\caption{Safety and performance trade-off: (a) slippery grid world with two options to satisfy formula (\ref{slippery_task}), where labelling is yellow: $\mathit{target}$, red: $\mathit{unsafe}$, blue: $\mathit{safe}$, and green is the initial state $s_0$; (b)~value function $V(s)$ without safe padding; and (c)~value function with safe padding (cautious RL).}%
	\label{fig_bridge_2}
\end{figure}

Nevertheless, the infinite number of state (and action) visits is one of the theoretical assumptions of QL asymptotic convergence~\cite{watkins}, which aligns with the proposed rate of change of~$\kappa$ and~$H$. Owing to time-varying $\kappa$ and $H$, when the agent synthesizes its policy, a subset of $\mathcal{A}^\otimes$ is only available, e.g., in the greedy case:
$$
a^*=\argmax_{a \in\mathcal{A}_p^H[1:\kappa]}~Q(s^\otimes,a)-r_p U_H(s^\otimes,a), 
$$ 
where the role of $r_p$ is to balance $Q$ and $U_H$. Note that since QL is an off-policy RL method, the choice of $a^*$ during learning does not affect the convergence of the Q-function \cite{watkins}. As the agent explores, the estimations of $P_a$, and thus of $U_H$, become more and more accurate, and the choice of actions become closer to optimal. Starting from its initial state $s_0$, the agent eventually expands the safe padding, i.e., the set of state-actions that it considers to be safe. The expansion occurs by diminishing the effect of the pessimistic learner, i.e., by decreasing the horizon $H$ of $U_H(s^\otimes,a)$ and also by increasing $\kappa$ in $\mathcal{A}_p^H(s^\otimes)$ until the effect of the pessimistic learner on decision making is minimal. Essentially, in the limit the role of the pessimistic agent is just to block the choice of actions that are critically unsafe according to $p_\mathit{critical}$ (actions that an optimal learned policy without the safe padding never takes, otherwise not optimal). However, the user-defined critical threshold $p_\mathit{critical}$ might affect the final policy of the agent in situations when acting safely may be at odds with acting optimally (Fig.~\ref{fig_bridge_2}). 

\begin{figure}[!t]
	\centering
	\vspace{-4mm}
	\subfloat[]{\includegraphics[width=0.8\linewidth]{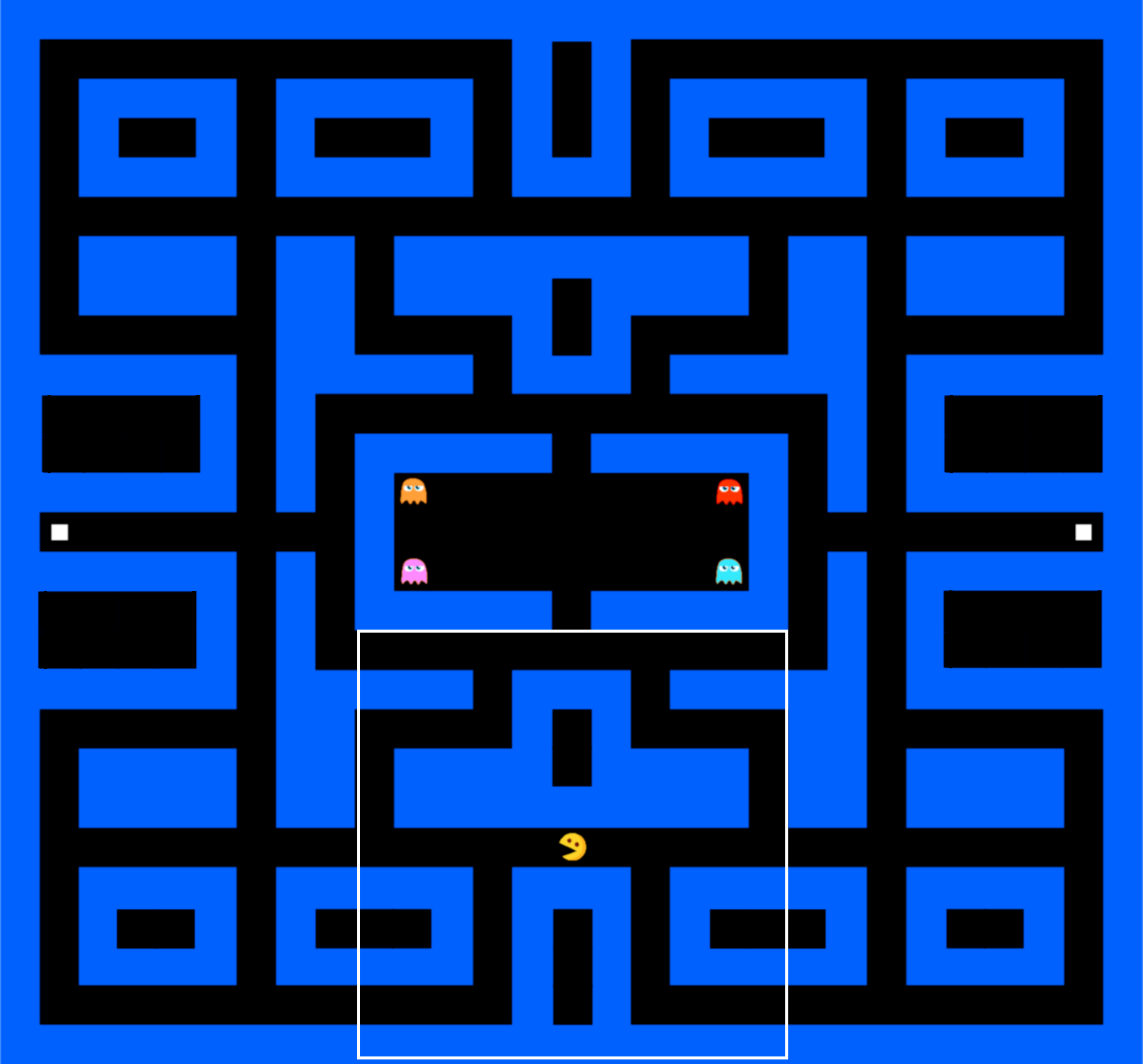}}
	\qquad
	\subfloat[]{
	\scalebox{0.8}{
	\begin{tikzpicture}[shorten >=1pt,node distance=2.3cm,on grid,auto] 
	\node[state,initial] (q_1) {$q_0$}; 
	\node[state] (q_2) [above right=of q_1] {$q_1$}; 
	\node[state] (q_3) [below right=of q_1] {$q_2$}; 
	\node[state] (q_5) [right=of q_1] {$q_4$}; 
	\node[state,accepting] (q_4) [right=of q_5] {$q_3$}; 
	\path[->] 
	(q_1) edge [loop below] node {$n$} ()   	
	(q_1) edge [bend left=15] node {$f_1$} (q_2)
	(q_1) edge [bend right=15] node {$f_2$} (q_3)
	(q_2) edge [loop above] node[left,xshift=-0.1cm,yshift=-0.2cm] {$n \vee f_1$} ()
	(q_2) edge [bend right=-15] node {$f_2$} (q_4)
	(q_3) edge [loop below] node[right,xshift=0.1cm,yshift=0.2cm] {~~$n \vee f_2$} ()
	(q_3) edge [bend right=15] node {$f_1$} (q_4) 
	(q_2) edge node {$g$} (q_5)
	(q_3) edge node {$g$} (q_5)
	(q_1) edge node {$g$} (q_5)
	(q_5) edge [loop right] node {$\mathit{true}$} ()
	(q_4) edge [loop above] node {$\mathit{true}$} ();
	\end{tikzpicture}}
	}\\*
	\subfloat[]{\hspace{-3mm}\includegraphics[width=0.55\linewidth]{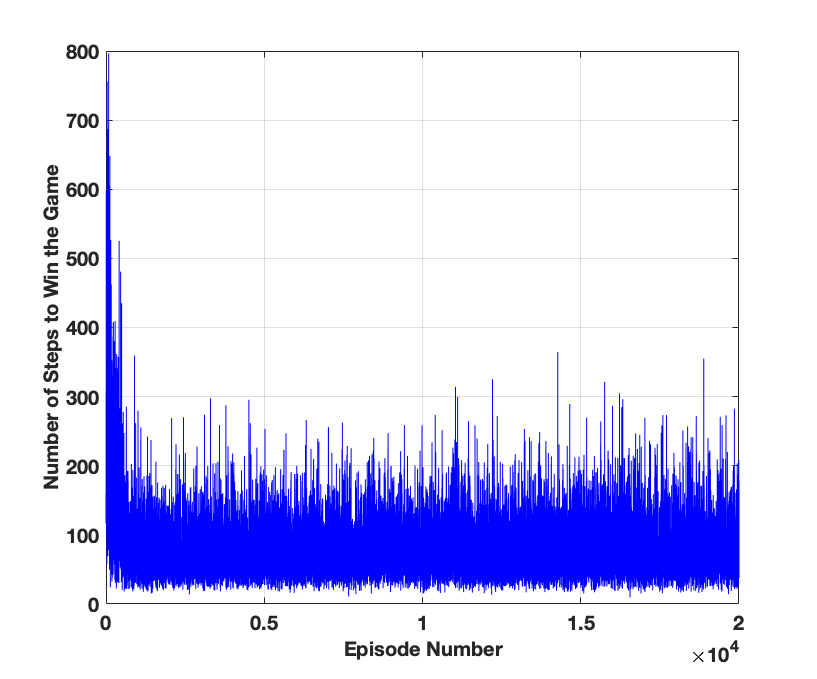}}
	\scalebox{0.95}[1]{\subfloat[]{\hspace{-3mm}\includegraphics[width=0.593\linewidth]{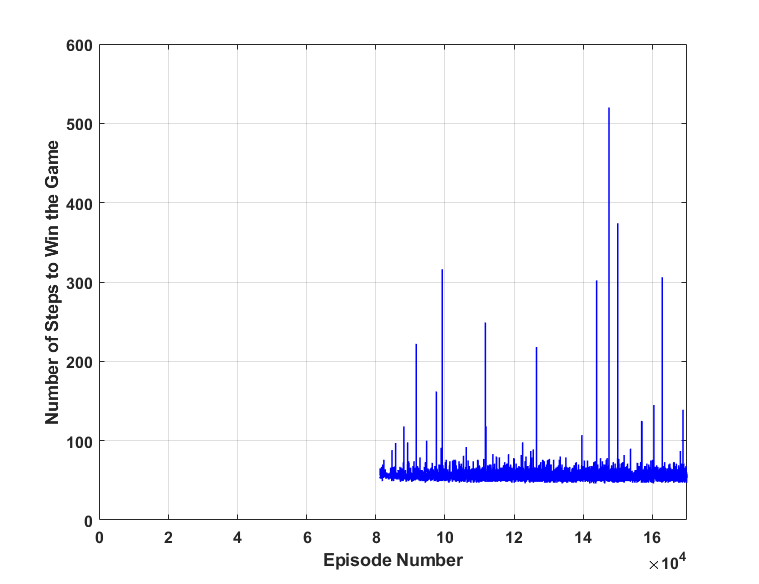}}}
	\caption{(a) Pacman environment with $|\mathcal{S}| > 80000$.  Observation area is large square around initial condition. Square on the left is labelled as food~1 ($ f_1 $) and that the one on the right as food~2 ($ f_2 $), the state of being caught by a ghost is labelled as $ g $, and the rest of the state space is neutral ($ n $); (b) LDBA for the specification (\ref{pacman_p}); (c) number of steps to complete the game with safe padding on (cautious RL), (d) and with safe padding off.}
	\label{pacmaninit} 
\end{figure}
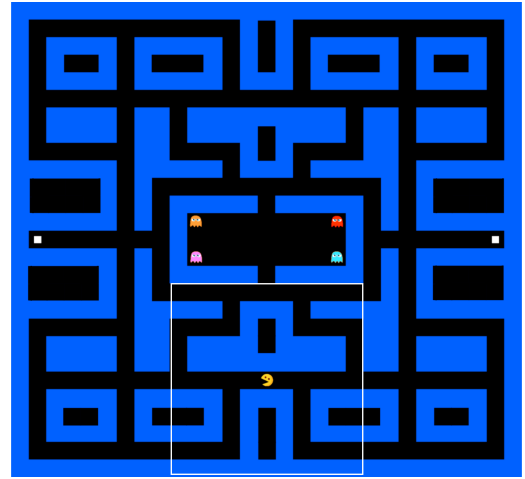
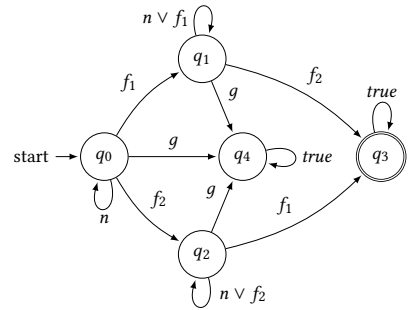
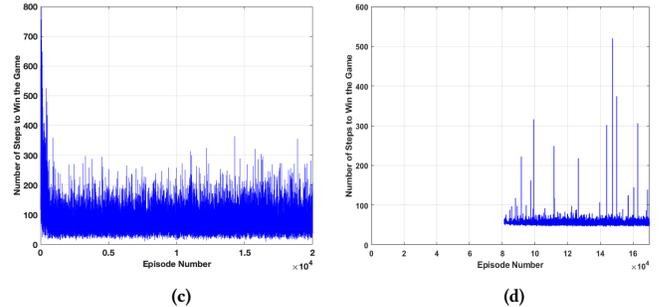

\section{Experiments}\label{case study}

We consider numerical experiments that concern LTL-constrained safe control policy synthesis problems for a robot in a slippery grid-world and the classical Pacman game. In both experiments the agent has to experience risky situations in order to achieve the goal. This allows us evaluate the performance of the proposed safe padding architecture in protecting the agent from entering unsafe states. 

For the robot example, let the grid be a $ 20 \times 20 $ square over which the robot moves. In this setup, the robot location is the MDP state $s \in \mathcal{S} $. At each state $s \in \mathcal{S}$ the robot has a set of actions $ \mathcal{A}=\{\mathit{left},\mathit{right},\mathit{up},\mathit{down},\mathit{stay}\}$ using which the robot is able to move to other states (e.g. $s'$) with the probability of $P(s,a,s'), a \in \mathcal{A}$. At each state $s \in \mathcal{S}$, the actions available to the robot are either to move to a neighbour state $s' \in \mathcal{S}$ or to stay at the state $s$. In this example, we assume for each action the robot chooses, there is a probability of $85\%$ that the action takes the robot to the correct state and $15\%$  that the action takes the robot to a random state in its neighbourhood, including its current state.

To get to the target state the agent has to cross a bridge (Fig.~\ref{fig_bridge}a) surrounded by unsafe states. The grid is slippery, namely from the agent's perspective, when it takes an action it usually moves to the intended cell, but there is an \emph{unknown} probability that the agent is moved to a random neighbour cell. However, the agent prior belief $P_a$ is that it can always move to the correct state and this is the dynamics known to the agent. The initial state of the agent is bottom left, $\gamma=0.9$, $\mu=0.85$, $p_\mathit{critical}=0.82$, and the observation radius is $r_o=2$. Note that for the sake of exposition, we intentionally picked $p_\mathit{critical}=0.82$ close to the grid-world slipperiness probability of $0.85$ and $r_o$ close to the bridge gap, while in practice when the environment is unknown, $p_\mathit{critical}$ and $r_o$ should be set conservatively.

Similar to the pilot-helicopter example, the final goal of reaching the target is initially conflicting with the agent being safe since crossing the bridge has a high risk of slipping into an unsafe state. Thus, the agent has to slowly try different states while remaining safe, until it realises that there is no other way than crossing the high-risk bridge to achieve its goal. The LTL property associated with this task is as follows: 
\vspace{-1mm}
\begin{equation}
	\label{slippery_task}
	\lozenge \mathit{target} \wedge \square \neg \mathit{unsafe}. 
\end{equation}

Notice that in this example the safety requirements we uphold while learning are embedded directly within the LTL formula for the task. In general the two requirements can be distinct. 

To ensure the agent's safety, we create a safe padding based on the agent knowledge of its own dynamics. This safe padding gradually grows, allowing the agent to safely explore the MDP (Fig.~\ref{fig_bridge}c--d) while repelling the agent to get too close to unsafe areas. Thanks to this guarding effect of the safe padding, once the goal is reached, the agent can safely back-propagate the reward and shape the value function (Fig.~\ref{fig_bridge}b) according to which the safe policy is later generated. Furthermore, note that with Cautious RL the agent is focused on those parts of the state space that are most relevant to the satisfaction of the given LTL property.

There was no single incident of going to unsafe in this experiment even with such a limited observation radius (Table~\ref{success_stats}). With the safe padding on, training took 170 episodes for RL to converge and with safe padding off, it took 500 episodes. 

\begin{table}
	\centering
	\scalebox{0.9}{
	\begin{tabular}{|c|c|c|c|}
		\hline
		Case Study & Safe Padding & Fail Rate & Success Rate \\
		\hline \hline
		\multirow{2}{*}{Slippery Grid-world} & Off & 36.48\% & 63.52\% \\ \cline{2-4}
		& On & 0\% & 100\%
		\\ \hline \hline
		
		\multirow{2}{*}{Pacman} & Off & 52.69\% & 47.31\% \\ \cline{2-4}
		& On & 10.77\% & 89.23\%
		\\ \hline 
	\end{tabular}}
	\caption{Proportion of number of times that the agent ended in unsafe (fail) states, and proportion of number of times in which the agent finds a path satisfying the LTL specification during learning. Statistics are taken over 500 learning episodes in the slippery grid-world and over 20000 episodes in the Pacman experiment.}
	\label{success_stats}
\end{table}

The second experiment is the classic game Pacman, which is initialised in a tricky configuration likely to lead  the agent to be caught by the roaming ghosts  (Fig.~\ref{pacmaninit}a). In order to win the agent has to collect all tokens without being caught by ghosts: 
\begin{equation}
	\label{pacman_p}
	\lozenge [ (f_1 \wedge \lozenge f_2) \vee (f_2 \wedge \lozenge f_1)]  \wedge \square \neg g,  
\end{equation}
where the token on the left is labelled as $ f_1 $, the one on the right as $ f_2 $, and the state of being caught by a ghost is labelled as $ g $. The constructed LDBA is shown in Fig.~\ref{pacmaninit}b. The ghost dynamics is stochastic: with a probability $ p_g=0.9 $ each ghost is chasing Pacman (\emph{chase mode}), else it executes a random move (\emph{scatter mode}). Note that each combination of (Pacman, ghost1, ghost2, ghost3, ghost4) represents a state in the experiment, resulting in a state-space cardinality over $80000$. As in the previous case study, here the safety requirements we uphold while learning are embedded directly within the LTL formula for the task.  

Fig.~\ref{pacmaninit}c gives the results of learning with safe padding on and Fig.~\ref{pacmaninit}d off. Note that with the safe padding on, the agent was able to successfully escape the ghosts even from the beginning, with the cost of longer path to win whereas, without the safe padding it took $80000$ episodes to score the very first win. In the Pacman experiment, the safe padding significantly reduced the number of times the agent got caught by the ghosts (Table~\ref{success_stats}).

\section{Conclusions}

In this paper, we have proposed \emph{Cautious Reinforcement Learning}, a~general method for safe exploration in RL usable on black-box MDPs, which ensures agent safety both during the learning process and for the final, trained agent. 
The proposed safe learning approach is in principle applicable to any standard reward-based RL. We have employed Linear Temporal Logic (LTL) to express an overall task (or goal), and to shape the reward for the agent in a provably-correct and safe scheme. 
We have proposed a double-agent RL architecture: one agent is pessimistic and limits the selection of the actions of the other agent, i.e., the optimistic one, which learns a policy that satisfies the LTL requirement. 
The pessimistic agent creates a  continuously growing ``safe padding'' for the optimistic agent, which can learn the optimal task-satisfying policy, while staying safe during learning. 
The algorithm automatically manages the trade-off between the need for safety during the training and the need to explore the environment to achieve the LTL objective. 

\subsubsection*{Acknowledgments}
{\footnotesize This work is in part supported by the HiClass project (113213), a partnership between the Aerospace Technology Institute (ATI), Department for Business, Energy \& Industrial Strategy (BEIS) and Innovate UK.}

\bibliographystyle{ACM-Ref}
\bibliography{Refs}
\end{document}